\documentclass[letterpaper, 10 pt, conference]{ieeeconf}  %
 
\usepackage{amsthm}

\usepackage{amsmath,amsfonts,bm}

\def\eqref#1{equation~\ref{#1}}

\def\1{\bm{1}}

\def\va{{\mathbf{a}}}

\def\vs{{\mathbf{s}}}

\DeclareMathAlphabet{\mathsfit}{\encodingdefault}{\sfdefault}{m}{sl}
\SetMathAlphabet{\mathsfit}{bold}{\encodingdefault}{\sfdefault}{bx}{n}

\def\gN{{\mathcal{N}}}

\newcommand{\tnll}{\mathcal{L}_{\text{NLL}}\xspace}
\newcommand{\tce}{\mathcal{L}_{\text{CE}}\xspace}
\newcommand{\tdc}{\mathcal{L}_{\text{D}}\xspace}
\newcommand{\tgen}{\mathcal{L}_{\text{G}}\xspace}
\newcommand{\treal}{\mathcal{L}_{\text{real}}\xspace}
\newcommand{\tfake}{\mathcal{L}_{\text{fake}}\xspace}
\newcommand{\tmarg}{\rho_{\pi_\theta}}
\newcommand{\tpol}{{\pi_\theta}}
\newcommand{\tll}{\mathcal{L}_\pi(\theta)\xspace}
\newcommand{\tdd}{\mathcal{L}_{D}(\phi)\xspace}

\newcommand{\vrtg}{\mathbf{\hat{R}}}
\newcommand{\Tau}{\mathcal{T}}

\DeclareMathOperator*{\argmin}{arg\,min}

\usepackage[breaklinks=true,colorlinks,citecolor=brown]{hyperref}
\usepackage{url}
\usepackage{comment}
\usepackage{makecell, multirow}
\usepackage{multirow}
\usepackage{booktabs}
\usepackage{amsmath}
\usepackage{array}
\usepackage{mathrsfs}
\usepackage{amssymb}
\usepackage{amsfonts}
\usepackage{amsmath}
\usepackage[flushleft]{threeparttable}
\usepackage{tablefootnote}
\usepackage{multirow}
\usepackage{hhline}
\usepackage{graphicx}
\usepackage{graphicx, subcaption}
\usepackage{hanging}
\usepackage{color}
\usepackage{tikz}
\usepackage{float}
\usetikzlibrary{calc,arrows,decorations.markings}
\usepackage{xcolor}
\usepackage{hhline}
\usepackage{algorithm}
\usepackage{algpseudocode} 
\usepackage{ifthen}
\usepackage{xspace}
\usepackage[automake,toc,abbreviations]{glossaries-extra}
\setabbreviationstyle{long-short}
\glsdisablehyper
\newabbreviation{rl}{RL}{Reinforcement Learning}

\newabbreviation{rtg}{RTG}{reward-to-go}

\newabbreviation{mrac}{MRAC}{Model Reference Adaptive Control}

\newabbreviation{ddmpc}{DD-MPC}{Data-Driven MPC}

\newabbreviation{nlp}{NLP}{Nonlinear Program}

\newabbreviation{cl}{C.L.}{closed-loop}

\newabbreviation{ol}{O.L.}{open-loop}

\newabbreviation{qp}{QP}{Quadratic Program}

\newabbreviation{lti}{LTI}{linear time-invariant}

\newabbreviation{blr}{BLR}{Bayesian Linear Regression}

\newabbreviation{bnn}{BNN}{Bayesian Neural Network}

\newabbreviation{gmm}{GMM}{Gaussian Mixture Model}

\newabbreviation{gcrl}{GCRL}{Goal-Conditioned Reinforcement Learning}

\newabbreviation{mlp}{MLP}{Multi-Layer Perceptron}

\newabbreviation{dt}{DT}{Decision Transformer}

\newabbreviation{atamp}{ATAMP}{Adaptive Task Action Motion Planner}

\newabbreviation{gan}{GAN}{Generative Adversarial Network}

\newabbreviation{rvs}{RvS}{Reinforcement learning via Supervised Learning}

\newabbreviation{bc}{BC}{behaviour cloning}

\newabbreviation{gc}{GC}{Goal-Conditioning}

\newabbreviation{mdp}{MDP}{Markov Decision Processes}
\newabbreviation{ga}{GA}{Goal Augmentation}

\makeglossaries
\usepackage{thmtools}
\declaretheorem[name=Example]{example}

\newcommand{\sect}[1]{Section~\ref{#1}}
\newcommand{\ssect}[1]{\S~\ref{#1}}
\newcommand{\eqn}[1]{Equation~\ref{#1}}
\newcommand{\fig}[1]{Figure~\ref{#1}}

\newcommand{\myparagraph}[1]{{\bf #1}\quad}

\newcommand{\model}{Adaptformer\xspace}

\definecolor{MyDarkBlue}{rgb}{0,0.08,1}
\definecolor{MyDarkGreen}{rgb}{0.02,0.6,0.02}
\definecolor{MyDarkRed}{rgb}{0.8,0.02,0.02}
\definecolor{MyDarkOrange}{rgb}{0.40,0.2,0.02}
\definecolor{MyPurple}{RGB}{111,0,255}
\definecolor{MyRed}{rgb}{1.0,0.0,0.0}
\definecolor{MyGold}{rgb}{0.75,0.6,0.12}
\definecolor{MyDarkgray}{rgb}{0.66, 0.66, 0.66}

\newcommand{\te}[1]{\texttt{#1}}
\newcommand{\disc}{\mathbf{D}_{\phi}}
\newcommand{\textclr}[1]{\textcolor{black}{#1}}

\newcommand{\approptoinn}[2]{\mathrel{\vcenter{
  \offinterlineskip\halign{\hfil$##$\cr
    #1\propto\cr\noalign{\kern2pt}#1\sim\cr\noalign{\kern-2pt}}}}}

\IEEEoverridecommandlockouts                              %

\title{\LARGE \bf
\model: Sequence models as adaptive iterative planners
}

\author{Akash Karthikeyan$^{1}$ and Yash Vardhan Pant$^{1}$%
\thanks{$^{1}$University of Waterloo,
        {\tt\small a9karthi@uwaterloo.ca}, {\tt\small yash.pant@uwaterloo.ca}}%
}

\begin{document}

\maketitle
\thispagestyle{empty}
\pagestyle{empty}

\begin{abstract}
Despite recent advances in learning-based behavioral planning for autonomous systems, decision-making in multi-task missions remains a challenging problem. For instance, a mission might require a robot to explore an unknown environment, locate the goals, and navigate to them, even if there are obstacles along the way. Such behavioral planning problems are difficult to solve due to: a) sparse rewards, meaning a reward signal is available only once all the tasks in a mission have been satisfied, and b) the agent having to perform tasks at run-time that are not covered in the training data (demonstrations), e.g., demonstrations only from an environment where all doors were unlocked. 
Consequently, state-of-the-art decision-making methods in such settings are limited to missions where the required tasks are well-represented in the training demonstrations and can be solved within a short (temporal) planning horizon.
To overcome these limitations, we propose \model, a stochastic and adaptive planner that utilizes sequence models for sample-efficient exploration and exploitation. This framework relies on learning an energy-based heuristic, which needs to be minimized over a sequence of high-level decisions. To generate successful action sequences for long-horizon missions, \model aims to achieve shorter sub-goals, which are proposed through an intrinsic (learned) sub-goal curriculum.
 Through these two key components, \model allows for generalization to out-of-distribution tasks and environments, i.e., missions that were not a part of the training data. Empirical results in multiple simulation environments demonstrate the effectiveness of our method. Notably, \model not only outperforms the state-of-the-art method by up to $25\%$ in multi-goal maze reachability tasks, but it also successfully adapts to multi-task missions that the state-of-the-art method could not complete while leveraging only demonstrations (for training) on single-goal-reaching tasks.\footnotemark
\footnotetext{\href{https://aku02.github.io/projects/adaptformer/}{https://aku02.github.io/projects/adaptformer/}}

\end{abstract}

\section{Introduction}
\begin{figure}[tb]
    \centering
    \includegraphics[width = 0.5\textwidth]{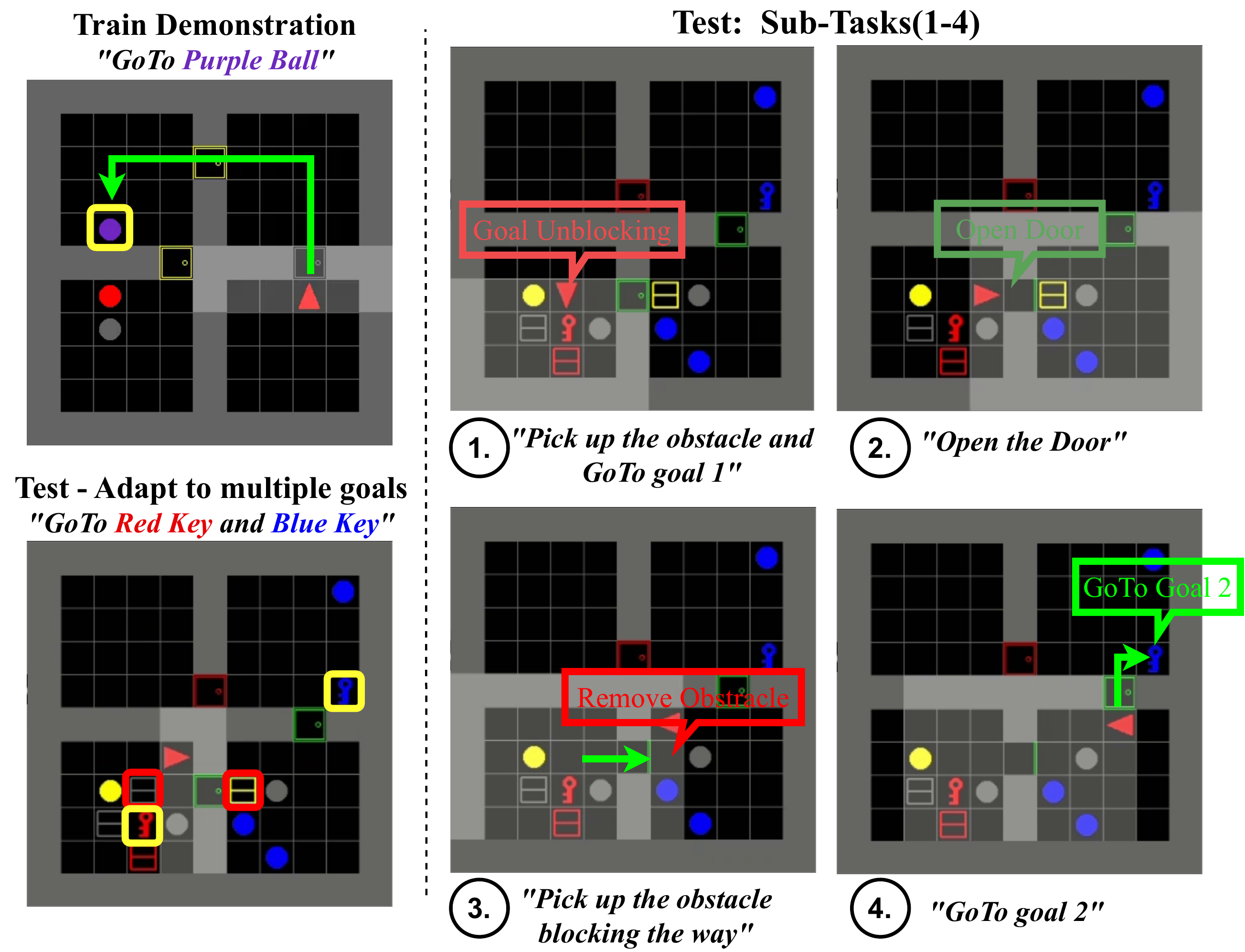}
    \caption{\small{\textbf{Multi-Task Mission Adaptation.} AdaptFormer plans a goal-conditioned trajectory addressing several key challenges: 1. recognizing and executing implicit subtasks ($1\rightarrow4$) in long-horizon missions, 2. generalizing to tasks involving multiple goals, and 3. adaptive skill learning (i.e., unblocking pathways) using an iterative stochastic policy. Goals are highlighted in yellow, while distractors are marked in red.}}
    \label{fig:teaser}
    \vspace{-12pt}
\end{figure}

An intelligent autonomous agent must be adaptable to new tasks at runtime, beyond those encountered during training. This is crucial for operating in complex environments that may introduce distractors (i.e., objects the agent has not seen before) and have multiple novel goals.
\begin{example}
\label{ex:adapt}
Consider the mission depicted in~\fig{fig:teaser}, where the agent navigates a complex environment. This environment is further complicated by the presence of additional distractors and the requirement to adapt to multiple goals during run-time. Note, that the goal position is not available to the planner, requiring exploratory actions to achieve the objectives. Furthermore, the presence of distractors necessitates adaptive actions, such as unblocking paths, to access doors and goal positions.
\end{example}

Recent approaches, such as \gls{rvs}, offer a simpler alternative to traditional \gls{rl} techniques by leveraging a \gls{bc} objective. \gls{rvs} effectively circumvents the complexities associated with Temporal Difference learning and avoids the ``deadly triad" of function approximation, bootstrapping, and off-policy learning~\cite{sutton2018reinforcement}, which can otherwise lead to poor performance in apriori unseen environments \cite{PENG2024111341}. Sequence models, a subset of \gls{rvs}, utilize the \gls{bc} objective to overcome challenges such as long-term credit assignment. 
Their effectiveness, however, is highly dependent on the diversity of training data. Our empirical studies, as discussed in Section~\ref{sec:random_result}, reveal that state-of-the-art methods struggle with tasks that were not covered in the training demonstrations. Specifically, skills that are not demonstrated in the training data (\textit{i.e., door opening}, or \textit{obstacle unblocking}) remain unlearned.

Learning to perform missions with long horizons also remains a long-standing challenge. Recently, \gls{gc} \cite{ijcai2022p770} has shown promise in long-horizon planning, but does not generalize well to apriori unseen goals. 
Additionally, when the goal is far away from the agent's current position, \gls{gc} does not consistently provide clear guidance for learning a policy. This can result in stalling actions or the agent getting stuck in a loop. This inconsistency contributes to the failures observed in both \gls{dt}~\cite{chen_decision_2021} and LEAP~\cite{chen_planning_2023}, which we also observe in our simulation studies (see section \ref{sec:result}).  %

\noindent{\textbf{Contributions of this work.}}
Here, we develop a planning approach that addresses the aforementioned challenges. We introduce \model, an adaptive planner, which:
\begin{enumerate}
    \item is capable of long-horizon task planning by learning a generative intrinsic goal curriculum,
    \item learns a (stochastic) policy that is subject to an entropy-based constraint. This not only facilitates adaptability, allowing for generalization to previously unseen tasks and environments, but also enhances the overall planning capability.
\end{enumerate}

Through extensive simulations, we empirically demonstrate that AdaptFormer outperforms the state-of-the-art by up to $25\%$ in multi-goal maze reachability tasks. It achieves this by adapting to various multi-goal tasks using only single-goal-reaching demonstrations, without requiring additional demonstrations. Furthermore, we demonstrate that \model is capable of learning from sub-optimal (random) demonstrations, while LEAP~\cite{chen_planning_2023} struggles to do so.

\section{Related Works}

\begin{figure*}
    \centering
    \includegraphics[width=\textwidth, trim=0 70 0 0, clip]{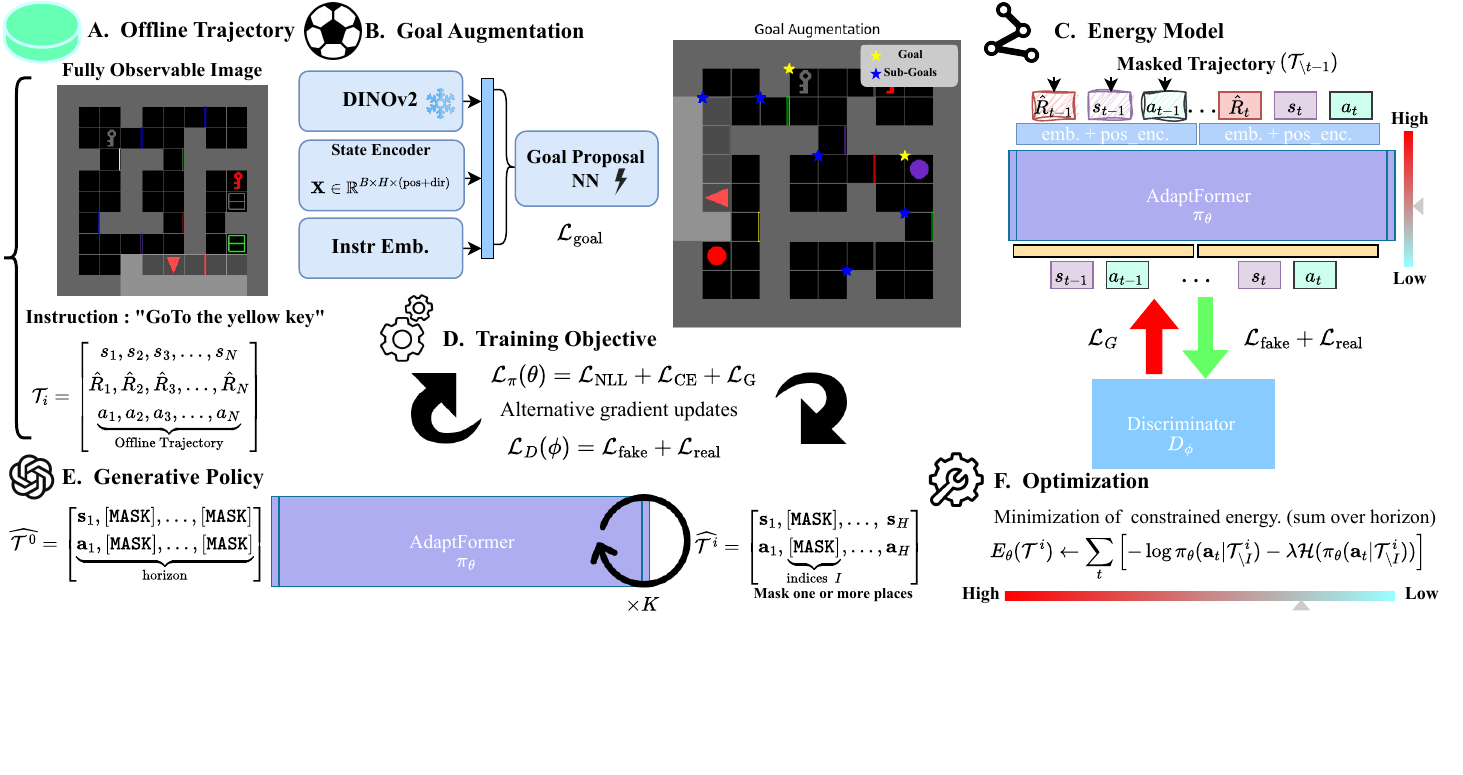} %
    \caption{{\small\textbf{Method Overview.} The \model, trained on offline data (\textbf{A}), incorporates a Goal Augmentation module that outputs a set of waypoints (\textbf{B}). Concurrently, the energy module is designed to assign lower energy to an optimal set of actions (\textbf{C}). Training involves alternating gradient updates to both the generator and the discriminator (\textbf{D}), promoting the policy to learn diverse representations. At the inference stage, the system employs the learned stochastic policy to query the masked trajectory sequence (\textbf{E}), which is then refined through iterative energy minimization (\textbf{F}), framing the path planning as an optimization problem.}}
    \label{fig:method}
\end{figure*}

\noindent\textbf{Offline \gls{rl}.} Offline \gls{rl} focuses on learning policies from collected datasets as seen in~\fig{fig:method}\textcolor{red}{A}, without any additional interaction with the environment~\cite{levine2020offline}. 
This approach faces the challenge of \textit{distribution shift} between the training demonstrations and run-time distribution. 
To address the distribution shift, several regularization strategies have been proposed. These include reducing the discrepancy between the learned and behavioral policies~\cite{fujimoto2019off, kumar2019stabilizing, fujimoto2021minimalist}, as well as implementing value regularization to impose implicit constraints, such as optimizing policies based on conservative value estimations~\cite{kumar2020conservative}. Despite its potential for learning from training demonstrations, these approaches often encounter difficulties in adapting to new scenarios, as highlighted in Example~\ref{ex:adapt}. Offline methods generally employ a pessimistic approach to value function estimation~\cite{levine2020offline}, causing instability in training and poor generalization.

\noindent\textbf{Sequence Models in \gls{rl}.} Sequence models in deep learning have been extensively studied in the domain of language modeling, from early sequence-to-sequence models~\cite{LSTM1997} to BERT~\cite{devlin2018bert}.
In \gls{rl}, these sequence models have been applied to learn value functions and compute policy gradients, leading to improved performance through model architecture, such as convolutions and self-attention mechanisms, which enable temporally and spatially consistent predictions.
More recent works~\cite{chen_decision_2021, janner_offline_2021, furuta2021generalized} have adopted an autoregressive modeling objective. This approach leverages the conditional generative capabilities of sequence models, where conditioning on desired returns or goal states facilitates the generation of future actions leading to those states or returns, assuming they were observed during training. These strategies aim to answer the question of what action is typically taken next, based on experience, assuming that the desired outcome will occur~\cite{chen_decision_2021, paster_you_2022}. Such a behavior cloning approach is designed to map observations to actions, with guidance signals indicating the closeness of the agent's actions to those observed in demonstrations. While effective for behavior cloning tasks, these methods fail in scenarios as seen in Example~\ref{ex:adapt}.

\noindent\textbf{Planning with Sequence Models.} LEAP~\cite{chen_planning_2023} formulates planning as an iterative energy minimization problem, wherein trajectory-level energy functions are learned via a masked language modeling objective. This method demonstrates generalization to novel test-time scenarios. However, its efficacy is contingent upon an oracle for goal positions essential for generating goal-conditioned trajectories. In the absence of such an oracle, there is a notable decrease in performance, particularly evident in larger mazes, we observe the same in our simulation studies   (see Table~\ref{tbl:performance_comparison}). This dependence significantly reduces the model's ability to generalize apriori unseen goals. The major drawback of lack of such conditioning is that it can cause the agent to enter loops, inhibiting \textit{exploration} and leading to a vulnerability of \textit{stalling actions}—the agent is stuck at the current position without advancing to the next state. To address the above issues, \model learns an intrinsic, goal curriculum that generalizes to a diverse goal distribution, which allows it to adapt to apriori unseen goals at run-time. Additionally, adopting a stochastic policy allows for exploration and adaptive skill learning.

\noindent\myparagraph{Outline of the paper.} In \sect{sec:ps}, we formalize the objective we aim to solve. \sect{sec:method}, introduces the \model framework an it's components. The overall framework is presented in \fig{fig:method}. The training and planning processes are then outlined in Sections \ref{sec:training} and \ref{sec:plan}, respectively. We present extensive empirical results and discussions in \sect{sec:result}.

\section{Preliminaries and Problem Statement}
\label{sec:ps}
\subsection{\textbf{Goal-Augmented \gls{mdp}}} 
We extend the standard \gls{mdp} framework by incorporating a set of goals \(G\) within the state space \(S\). Formally, we consider learning in a Goal-augmented \gls{mdp} $\mathcal{M} = \langle S, A, T, G, R \rangle$, The MDP tuple comprises discrete states $s \in S$, each represented by a 2D position, direction and a fully observable image $(x, y, d, I)$ within a finite set of states;  discrete actions $a \in A$; the unknown transition dynamics $T(\cdot|s, a)$; a set of absorbing goal states $G \subset S$ and a delayed reward function $R: S \rightarrow \{0,1\}$~\cite{sutton2018reinforcement,pitis20a}. \textclr{Additionally, we have instructions (INS) that specify the goals of a particular task} (see figure~\ref{fig:method}\textcolor{red}{A}).  
We relabel the reward function \(R_g(s)\) to encourage the agent to seek shorter path sequences, defined as $R_g(s) = \mathbb{I}\{s \in G\} + c$, where \(c = -1\). The agent gets a reward of $1$ when the final goal is reached and $-1$ otherwise. The agent's objective, \(\mathcal{J}\), is to maximize the expected discounted cumulative reward through a goal-conditioned policy \(\pi: S\times G \rightarrow A\). The discount factor, \(\gamma \in [0,1)\), modulates the importance of immediate versus future rewards.
\begin{equation}
\label{eqn:base}
     \mathcal{J}(\pi) =  \mathbb{E}_{\substack{a_t \sim \pi(\cdot | s_t, g), \\ g \in G, \\ s_{t+1} \sim T(s_t, a_t)}} \left[\sum_{t=0}^{\infty} \gamma^t R_g(s_t)\right]
\end{equation}
Here, \(t\) corresponds to the timestep. The sparse, binary reward setting often leads to ambiguous guidance and can be uninformative \cite{pitis20a}. Moreover, we aim to address test-time adaptability, such as adapting to multiple goals (including unseen apriori) or facilitating long-horizon planning. Therefore, we adopt an \gls{rvs}-based approach \cite{emmons2021rvs, chen_decision_2021}, extending policy learning as a conditional generator. To overcome these challenges, we propose to formulate \eqref{eqn:base} as a \gls{rvs} learning objective, which we aim to optimize via offline \gls{rl}.

\subsection{\textbf{Offline \gls{rl}}}
\label{sec:offline_rl}
In offline \gls{rl}, we have access to a dataset of near-optimal trajectories collected through demonstrations in the environment as seen in~\fig{fig:method}\textcolor{red}{A}. A trajectory is a sequence of states, actions, and return-to-go $\vrtg_t = \sum_{k=t}^{n} R_g(s_k)$, with length $n$: $\Tau_N = \left(s_1, a_1, \vrtg_1, \ldots, s_N, a_N, \vrtg_N\right)$. The dataset consists of multiple such trajectories. 
To solve the policy learning problem via offline RL, we use
masked language model to learn energy functions, which allow us to reframe the planning in \eqref{eqn:base} as an iterative optimization procedure~\cite{chen_planning_2023}. Here, a masked language model, parameterized by \(\theta\), is trained to learn a locally normalized energy function \(E_\theta\). 
This function assigns a scalar value to each trajectory, such that an optimal sequence of actions \(\va_{1:T}^*\) conditioned on the goals \(G\), receives a low energy score. The policy aims to identify a sequence of actions for a task that minimizes this energy, learned based on a set of offline demonstrations. Here \(H\) is a hyper-parameter, which denotes the planning horizon.
\begin{equation}
\label{eq:overall}
\va_{1:H}^* 
= \argmin_{\va_{1:H}}  E_{\theta} ( \va_{1:H}|\Tau_{\textbf{H}}, G)
\end{equation}
\label{sec:ps_challenges}
\noindent\textbf{Challenges.} While the objective in~\eqn{eq:overall} facilitates learning an energy function that acts as a surrogate to rewards \eqref{eqn:base}, in turn enabling generation of goal-conditioned trajectories, several challenges remain: (1) Goals at test time could be unknown to the agent\footnote{In this work, we consider the case when the goals at test time are apriori unknown. See section \ref{sec:result} for simulation results.}, (2) Even when goals are specified, goal-conditioned trajectories may offer insufficient guidance for long-horizon tasks, especially when the goal is far from the agent's current position \cite{badrinath_waypoint_2023}, (3) The agent may fail to adapt to multiple goals at test time, potentially hallucinating trajectories observed in training (i.e., generates trajectories that are similar to those seen in demonstrations), (4) Certain skills required for the mission remain unlearned if they are not observed in the training demonstrations (\textit{i.e., door opening}, or \textit{obstacle unblocking}).
\section{Adaptformer: Methodology}
\label{sec:method}
\subsection{\textbf{Overview of \model}}
\label{sec:sub_ps}
To address the challenges discussed in Section~\ref{sec:ps_challenges}, we propose \emph{Adaptformer}. The framework consists of the following key components: a \emph{Goal-Augmentation Module} to estimate the task-specific goal distribution at test time, discussed in~\sect{sec:ga}; a \emph{State Discriminator} that facilitates learning sample-efficient sub-goals to aid planning with long horizon tasks, detailed in~\sect{sec:disc}; and an \emph{Energy based model} for an adaptive and generalizable policy in~\sect{sec:energy}. Section \ref{sec:training} describes the training of this model, and Section \ref{sec:plan} covers how the model is used for adaptive planning at run-time. 

\subsection{\textbf{\gls{ga}}}
\label{sec:ga}
Adaptformer aims to generate goal-conditioned trajectories. In contrast to existing work, e.g., \cite{chen_planning_2023, chen_decision_2021}, which relies on an oracle to obtain goal positions during online adaptation, we learn an intrinsic goal curriculum to achieve this. The \gls{ga} module learns a likelihood distribution across the state space and generates a set of goal proposals, denoted by $\hat{G}$. Its objective is defined as follows.
\begin{equation}
\max_{\omega} \log P(\hat{G} | G_{train} ; \pi_\omega)
\end{equation}
where the goal samples $\{g_i\}_{i =1}^N \in G_{train} \sim \Tau$ are drawn from the offline demonstrations. We parameterize the \gls{ga} module using a fully connected \gls{mlp} with parameters $\omega$, which accepts state, instruction, and image embedding as input~\cite{oquab2023dinov2} as seen in~\fig{fig:method}\textcolor{red}{B}, to generate goal proposals \(\hat{G}\). In practice, we upper bound the number of expected goal states, enabling \model to adapt to scenarios with multiple goal positions, even if it was initially trained on tasks with single goal. 

\subsection{\textbf{State Discriminator}}
\label{sec:disc}
We observe that guidance from goal-conditioned trajectories is often insufficient for long-horizon tasks, particularly when there is a distributional shift between training and test goal distributions, i.e., \(P(G_{train}) \neq P(G_{test})\). For instance, online adaptation might involve multi-goal missions, whereas \(G_{train}\) (demonstrations) primarily covers tasks with single-goal tasks. To address these challenges, we introduce learning sub-goal states \(S_g \subset S\),%
using a state discriminator~\cite{goodfellow2014generative, yang2023model}. The discriminator assigns confidence values in the range \((0,1)\) to assess whether the generated \textclr{samples of state sequences $\hat{\vs}_{1:H}$ are indistinguishable from the real sequence $\vs_{1:H}$. Consequently,
as the losses in~\eqn{eq:disc_2} converge, the discriminator learns the true state distribution and forces the policy to generate state sequence that is diverse yet in-distribution to those in demonstrations. This helps in better generalization to new tasks.}
To achieve this, the discriminator $D$, modeled as a \gls{mlp} with parameters \(\phi\), learns real state distribution by minimizing $\treal$ and discriminating the synthetic state sequence from the policy-induced marginal $\tmarg$\footnote{As the policy $\tpol$ converges, it learns the marginal distribution of states} by minimizing $\tfake$ in~\eqn{eq:disc_2}. Both the discriminator and the generative policy are trained alternately, as is standard in \gls{gan}.
Similar to \cite{badrinath_waypoint_2023}, we find that conditioning actions on achievable sub-goals are more conducive to generating optimal action sequences and preventing stalling. This also results in learning a sample-efficient goal curriculum. Since the discriminator is jointly trained with the policy, unlike hierarchical \gls{rl} approaches \cite{badrinath_waypoint_2023}, we do not need to specify a number of sub-goal proposals apriori. 
Thus, during inference, the policy can generate sample-efficient, intrinsic sub-goals as seen in~\fig{fig:method}\textcolor{red}{B}.
\subsection{\textbf{Energy Based Models for Trajectory Generation}}
\label{sec:energy}
Once the sub-goals have been established, they facilitate goal-conditioned policy learning. Given demonstrations (see section \ref{sec:offline_rl}), we aim to learn the energy of a trajectory \(\textit{E}_\theta(\Tau) \), defined as the sum of negative pseudo-likelihood over the horizon \(\textit{E}_\theta(\Tau) = \sum_{{t}=1}^{{H}} \left[ -\log \pi_\theta(\va_t | \Tau_{\backslash t}, S_g) \right]\)~\cite{goyal2021exposing}. This energy function assigns lower energy to an optimal sequence of actions as seen in~\fig{fig:method}\textcolor{red}{C}. Our approach leverages the masked language model objective \cite{goyal2021exposing} to learn a locally normalized energy score, allowing us to score generated rollouts and frame planning as an iterative optimization process~\cite{chen_planning_2023, goyal2021exposing}. We model a conditional generative policy \(\tpol\), optimized subject to lower bound \(\beta\) on its entropy \(\mathcal{H}\). This constraint encourages stochasticity and enhances the policy's adaptability to novel environments and tasks.
\begin{equation}
\label{eq:policy}
\begin{split}
\min_\theta \sum_{{t}=1}^{{H}} \left[ -\log \pi_\theta(\va_t | \Tau_{\backslash t}, S_g) \right] , \quad \text{s.t.} \\ \quad  \mathbb{E}_{\sim \Tau} \left[ \sum_{{t}=1}^{{H}} \mathcal{H}(\pi_{\theta}(\va_t | \Tau_{\backslash t}, S_g)) \right] \geq \beta
\end{split}
\end{equation}
Here \({H}\) denotes the planning horizon.
This formulation enables the learning of a stochastic policy \(\pi_\theta\) for action prediction at any given timestep. Unlike auto-regressive objectives, the masked language model objective incorporates a bidirectional context of actions across all timesteps, accounting for future trajectories \cite{devlin2018bert}.
Our framework thus emphasizes energy minimization across the entire planning horizon rather than focusing on individual timesteps. 
Next, we define how the \model is trained.
\subsection{\textbf{Training Objective}} 
\label{sec:training}
The objectives outlined in~\ssect{sec:method} \textcolor{red}{B-D} can be grouped in two, based on the parameters ($\theta, \phi$) being optimized as seen in~\fig{fig:method}\textcolor{red}{C}. Note, in our implementation, the Goal-Augmentation is a sub-module of the policy, hence $\omega$ is contained in $\theta$. Algorithm \ref{algo:iterative} gives an overview of the training procedure. The training involves alternately minimizing two loss functions: $\tll$ for the policy, and $\tdc(\phi)$ for the discriminator, described below:

\begin{subequations}
\begin{align}
\label{eq:nll}
\mathcal{L_\pi(\theta)} = \overbrace{\mathbb{E}_{ \sim \Tau}[-\log \mathbf{\pi}_\theta(\va_t |\Tau_{{\backslash t}}, S_g)]}^{\tnll}  \\
\label{eq:ce}
- \lambda_1 \overbrace{\mathbb{E}_{ \sim \Tau} [\mathcal{H}(\mathbf{\pi}_\theta(\cdot | \Tau_{\backslash t}, S_g))]}^{\tce}  \\
\label{eq:gen}
+ \lambda_2 \overbrace{\mathbb{E}_{(\hat{S})\sim \rho_{\pi_\theta}}[\log (1 - D_\phi(\hat{{{S}}}))]}^{\tgen} 
\end{align}
\end{subequations}
\begin{align}
\label{eq:disc_2}
\tdc(\phi)=\overbrace{\mathbb{E}_{(S)\sim {\Tau}}[\log D_\phi(S)]}^{\treal} \nonumber\\
+ \overbrace{\mathbb{E}_{(\hat{S})\sim \rho_{\pi_\theta}}[\log (1 - D_\phi(\hat{S}))]}^{\tfake} 
\end{align}
Note that the \eqn{eq:policy} can be transformed to the dual form, by introducing a Lagrangian multiplier $\lambda_1 \in (0, \infty]$. We perform alternate gradient decent steps on $\lambda_1$ and the $\tpol$. In practice, we observe that $\lambda_1 \rightarrow 0$ and the lower bound $\beta$ on entropy is eventually satisfied (similar to~\cite{zheng2022online}).
\begin{algorithm}[H]
	\caption{\model : Training Progression.} 
	\label{algo:iterative}
	\begin{algorithmic}[1]
	\State Initialize policy $\pi_\theta$, initialize discriminator $\disc$, offline demonstrations $\bm{{\Tau}}$, planning horizon $H$, learning rate $\alpha$
    \For {$t=1,\ldots, {H}$}
        \State \small{\color{gray}\te{// Mask $\va_t$, $\vs_t$, and $\vrtg_t$}}
        \State $\tnll^t \leftarrow \mathbb{E}_{({\va_t})\sim \Tau}[-\mathbf{\log \pi_\theta({\va_t}|\Tau_{\backslash t}, S_g)}]$ 
        
        \small{\color{gray}\te{// ${\va_t} \sim \gN(\mu_\theta(\Tau_{\backslash t}, S_g), \Sigma_\theta(\Tau_{\backslash t}, S_g))$}}
        \State $\tce^t$ $\leftarrow$ $\mathbb{E}_{(\va_t) \sim \Tau} [\mathcal{H}(\mathbf{\pi}_\theta(\va_t | \Tau_{\backslash t}, S_g))]$ 
        \State $\tgen^t$ $\leftarrow$ $\mathbb{E}_{(\hat{S})\sim \rho_{\pi_\theta}}[\log (1 - \disc(\hat{S}))]$ 
        \State $\tdc^t$ $\leftarrow$ $\mathbb{E}_{(S)\sim {\Tau}}[\log D_\phi(S)] + \mathbb{E}_{(\hat{S})\sim \rho_{\pi_\theta}}[\log (1 - D_\phi(\hat{S}))]$
        
    \EndFor
    \State ${\tll = \sum_{{t}=1}^{{H}} \tnll^t - \lambda_1 \tce^t + \lambda_2 \tgen^t}$ 
    \State $\theta \leftarrow  \theta - \alpha \nabla_\theta \tll$ 
    \State $\tdd = \sum_t \tdc^t $
    \State $\phi \leftarrow \phi - \alpha \nabla_\phi \tdd$ 
    \State $\lambda_1 \leftarrow \lambda_1 - \alpha (\mathcal{H}[\mathbf{\pi}_\theta(\cdot | \Tau_{\mathbf{\backslash t}})] - \beta)$ 
	\end{algorithmic} 
\end{algorithm}

\noindent\myparagraph{Negative Log-Likelihood (\(\tnll\)).} It corresponds to the loss for energy model described in~\sect{sec:energy}. The energy is defined as the summation of pseudo-likelihood values evaluated post-softmax. We adopt the same training strategy as seen in~\cite{devlin2018bert, chen_planning_2023} %

\noindent\myparagraph{Cross Entropy (\(\tce\)).} %
We incorporate the Shannon entropy regularizer \(\mathcal{H}(\tpol(\cdot | \Tau_{\backslash t}, S_g))\)~\cite{Haarnoja2018SoftAA}. Unlike approaches in MaxEnt \gls{rl} and SAC~\cite{haarnoja2018soft, Haarnoja2018SoftAA}, \model focuses on learning the trajectory-level energy across a planning horizon denoted by $H$, through the objective seen in~\eqn{eq:policy}. 
We design the policy to follow a gaussian distribution, with diagonal covariances.
This approach allows us to model a distribution over actions while also allowing us to model the covariances between different actions, in turn enabling us to compute $\mathcal{H}$. Here, the mean and log-variance are predicted by two separate fully connected \gls{mlp}s $\va \sim \gN(\mu_\theta(\Tau_{\backslash t}), \Sigma_\theta(\Tau_{\backslash t}))$~\cite{zheng2022online} (see line 4 in alg.~\ref{algo:iterative}).

\noindent\myparagraph{Generative Loss (\(\tgen\)).} It represents the guidance from discriminator, where the parameter $\lambda_2 \in (0,\infty]$, balances the accuracy and diversity of the learnt state distribution.
This approach prevents the policy from hallucinating trajectory sequences observed in training. By learning a diverse state representation, the generative loss enhances exploration capabilities. For instance, the model can predict sub-goals that were not part of the demonstrations. Consequently, the goal-conditioned policy can generate trajectories that traverse previously unseen regions of the environment during rollouts.

\noindent\myparagraph{Discriminator Loss (\(\tdc\)).} This loss function is utilized to update the discriminator's parameters seen in~\sect{sec:disc}, enabling it to adeptly differentiate between the generator's predictions and the true labels (line 7 in alg.~\ref{algo:iterative}).

Training proceeds alternatingly, with updates applied to the policy, the discriminator, and the temperature parameters, as outlined in Algorithm~\ref{algo:iterative}.
\subsection{\textbf{Planning: Online Adaptation.}} 
\label{sec:plan}
\begin{algorithm}[tb]
	\caption{\model: Rollout and Gibbs Sampling.} 
	\label{algo:iterative_plan}
	\begin{algorithmic}[1]
	\State \textbf{Require} Trained policy $\pi_\theta$, Plan Horizon $H$, current state $\vs_1$, mask token $\texttt{[M]}$
	\State Initialize \small{$\hat{\Tau}^0 = (\vs_1,\texttt{[M]}, \dots, \texttt{[M]}, \texttt{[M]})$}
     \State $E_\theta(\Tau^i) \leftarrow \sum_t[- \log \pi_\theta(\va_t|\Tau^{i}_{\backslash I}, S_g) - \lambda_1 \mathcal{H}(\pi_{\theta}(\mathbf{a}_t | \Tau^{i}_{\backslash I}, S_g))]$
    \For {$i=1,\ldots, K$} \small{\color{gray}\te{// K hyperparameter}}
        \State 
        $I \sim [1, 2, \cdots, H]$ \small{\color{gray}\te{// select indices $I$}}
        \State $\va, \vs \sim E_\theta(\Tau_{\backslash I}^i)$ \small{\color{gray}\te{// Query the energy model}}
        \State ${\Tau^{i+1}} \leftarrow {\Tau^i_{\backslash I} \cup a, s}$ \small{\color{gray}} \small{\color{gray} \te{// Update and repeat}}
    \EndFor 
	\end{algorithmic} 
\end{algorithm}

Algorithm \ref{algo:iterative_plan} shows how the learnt policy is used at test-time for planning in an iterative manner. Given the trained policy $\tpol$, we utilize Gibbs sampling~\cite{chen_planning_2023} to generate plans at run-time. We query the policy to score alternative actions at masked timesteps $t \in \{0,\dotsc,H\}$. At each iteration $i$, we query the actions and states $\va_t, \vs_t \sim E_\theta(\Tau_{\backslash t}^i)$ (line 5 in alg.~\ref{algo:iterative_plan}) as seen in~\fig{fig:method}. The positions of the masks are randomized. This strategy allows for %
minimizing the learnt energy function while also balancing exploration. 
We highlight a few comments regarding the iterative planning:
\begin{enumerate}
    \item The iterative goal-conditioned sampler prevents the model from stalling.
    \item It facilitates trajectory optimization by considering future states.
    \item The sub-goal proposals are influenced by the agent's current location, enabling the dynamic update of new sub-goals that guide the agent.
    \item Provides adaptability to scenarios not encountered in the demonstrations as described in Example~\ref{ex:adapt}.

\end{enumerate}
As seen in the next section, these are validated via extensive simulation studies.

\section{Results and Discussion}
\label{sec:result}
\begin{figure*}[tb]
    \centering
    \includegraphics[width=\textwidth]{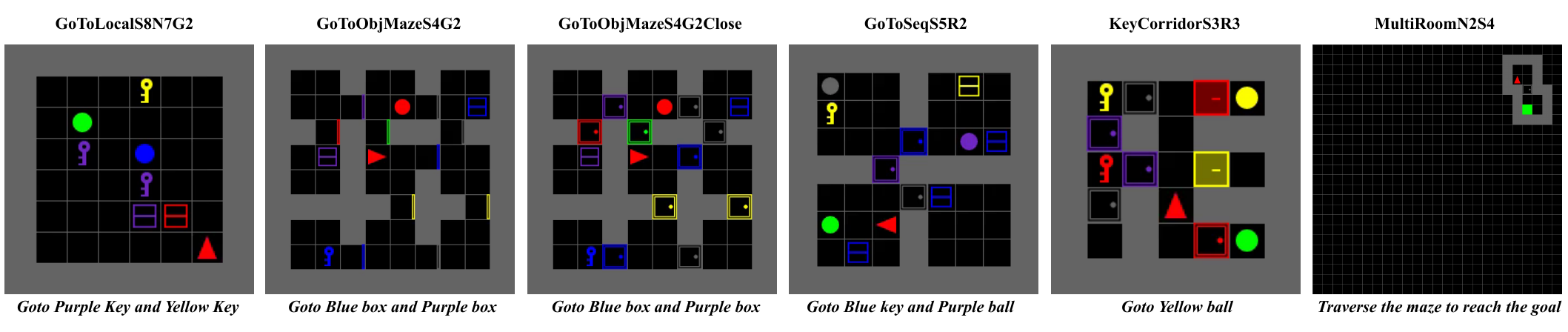} %
    \caption{{\small\textbf{Environments.} The simulations for both types of missions, (1) and (2) in section~\ref{sec:setup}, were conducted in the following mazes: the first five were sourced from BabyAI, while the final one was from MiniGrid.}}
    \label{fig:results}
\end{figure*}
\begin{figure}[h]
    \centering
    \includegraphics[width=0.4\textwidth]{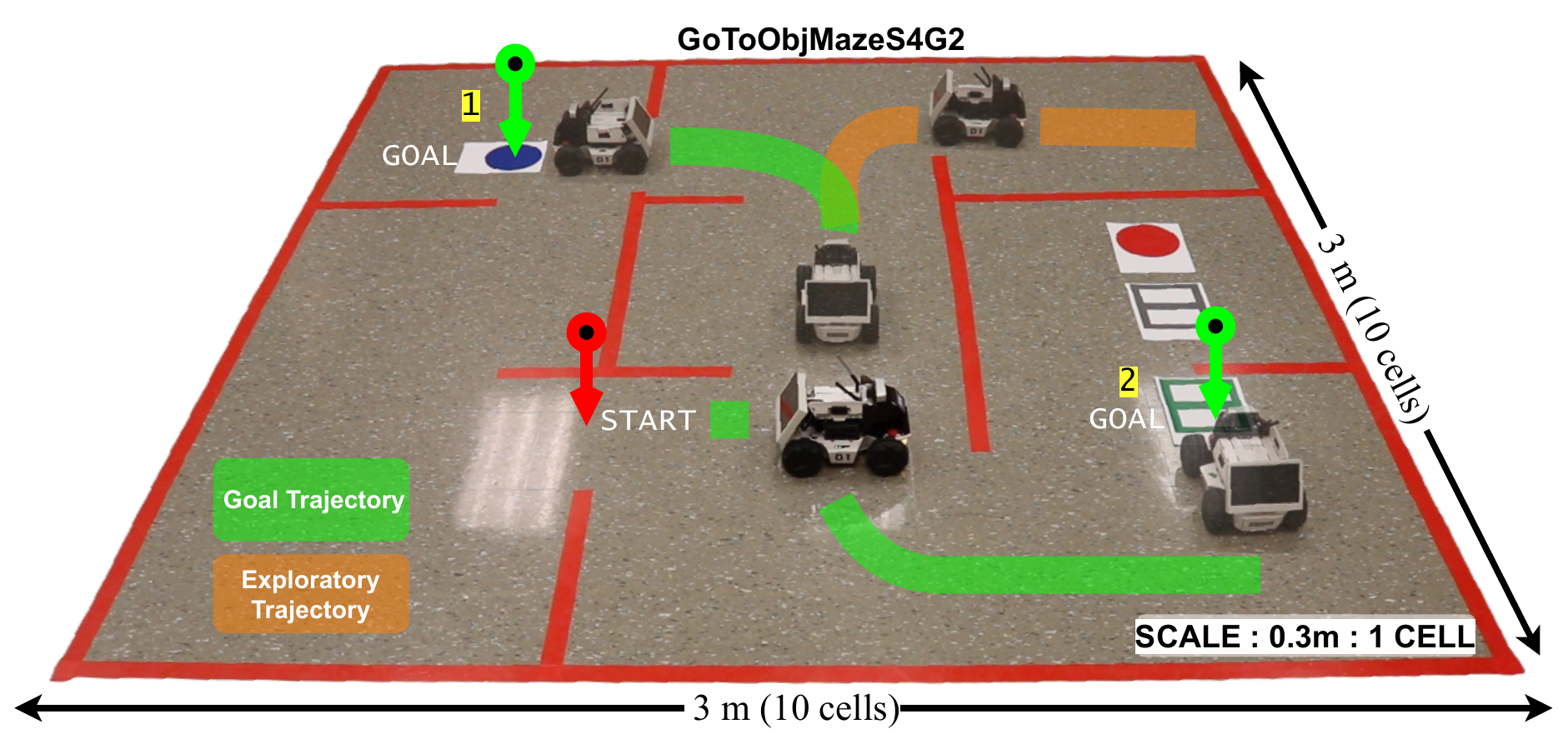}
    \caption{{\small\textbf{Policy Deployment.} We demonstrate the policy rollout in the AGILEX robot with a differential drive for the \textbf{\textit{GoToObjMazeS4G2}} task. Using a scale of 0.3 m per grid cell, and employing onboard odometry and dead-reckoning, the supplementary video showcases the implementation.}}
    \label{fig:bot}
    \vspace{-12pt}
\end{figure}

We assess the performance of our model in modified BABYAI~\cite{babyai_iclr19} and Minigrid~\cite{chevalierboisvert2023minigrid} environments, focusing on the following aspects:
\begin{enumerate}
    \item \textbf{Generalization.} We evaluate \model's effectiveness in trajectory planning within maze environments that were not a part of the demonstrations. %
    \item \textbf{Single-goal to Multi-goal Transfer (Adaptability).} The model, initially trained on single-goal-reaching demonstrations, leverages this acquired skill at test time to tackle multi-goal-reaching challenges. %
    \item \textbf{Auxiliary Distractors.} We introduce additional obstacles and goal states located across various rooms and behind closed doors. The obstacles are strategically placed around the goal state or the door region, requiring the agent to navigate around or reposition the distractor before reaching the goal state as seen in~\fig{fig:teaser}. Note that this goal unblocking behavior was not part of the training data (demonstrations).
\end{enumerate}
We implemented \model using {Python 3.8} and trained it on a 12-core CPU and an RTX A6000 GPU.  
We also deployed the policy on an {AGILEX} LimoBot (\fig{fig:bot}). 
See this webpage\footnotemark
\footnotetext{\href{https://aku02.github.io/projects/adaptformer/}{https://aku02.github.io/projects/adaptformer/}} for more details. 

\noindent\textbf{Baselines for comparison:} We compare \model with LEAP~\cite{chen_planning_2023} and its variant, LEAP without goal conditioning (LEAP$\ominus$GC) to evaluate the significance of conditioning in generalization to new maps. 
\subsection{\textbf{Simulation setup}} 
\label{sec:setup}
We run simulations over an extensive suite of tasks and various environments (\fig{fig:results}) across two paradigms (below). The agent can choose from among six actions, \textit{left, right, forward, open, drop, or pick up}, to navigate and interact with the environment. %
\begin{enumerate}
    \item \textbf{Trajectory Planning.} This involves the agent moving to one or more goals. %
    The agent is trained through a single goal-reaching task and then evaluated in a multi-goal environment. For our method and the baselines, we evaluate the success rates in reaching goal positions, reporting both the mean and variance. This evaluation is conducted across 50 maps, from 5 different starting positions for each map.

    \item \textbf{Instruction Completion.} Here, the agent operates in a multi-objective environment that will require a higher level of decision-making, such as \textit{exploration, picking/dropping objects, key collection, and then reaching the goal behind the door}. We evaluate success rates in reaching goal positions across 3 distinct seeds over 50 environment initializations (150 maps).
    
\end{enumerate}
\begin{table}[h!]
    \centering
    \color{black}
    \small
    \setlength\tabcolsep{4pt} %
    \renewcommand{\arraystretch}{1.2} %
    \begin{tabular}{lccc}
        \toprule
        \textbf{Environment} & \textbf{Ours} & \textbf{LEAP} & \textbf{LEAP$\ominus$GC} \\
        \midrule
        \multicolumn{4}{c}{\textbf{Trajectory Planning}} \\ 
        GoToLocalS8N7G2 & \textbf{98$\pm$1\%} & $88\pm1$\% & $92\pm2$\% \\
        GoToObjMazeS4G2 & \textbf{53$\pm$16\%} & $37\pm$29\% & $32\pm$30\% \\ 
        GoToObjMazeS4G2Close & \textbf{48$\pm$18\%} & $23\pm$20\% & $18\pm$20\% \\
        GoToObjMazeS4G1 & \textbf{64$\pm$18\%} & $58\pm$26\% & $54\pm$28\% \\ 
        
        \multicolumn{4}{c}{\textbf{Instruction Completion}} \\ 
        GoToSeqS5R2 & \textbf{45$\pm$5\%} & $43\pm$4\% & $39\pm$4\%\\
        KeyCorridorS3R3 & $18\pm$1\% & \textbf{21$\pm$3\%} & $16\pm$2\% \\
        
        \multicolumn{4}{c}{\textbf{Stochastic Environments}} \\  
        GoToLocalS8N7G2 & \textbf{98$\pm$1\%} & $95\pm$1\% & $97\pm$1\% \\
        GoToObjMazeS4G2 & \textbf{40$\pm$19\%} & $26\pm$20\% & $23\pm$20\% \\ 
        
        \multicolumn{4}{c}{\textbf{Randomly Collected Trajectory}} \\  
        MultiRoomN2S4 & \textbf{71\%} & 0\% & 0\% \\
        \bottomrule
    \end{tabular}
    \caption{{\small\textbf{Quantitative performance comparison of our model and LEAP.} Success rates of the models across different environments are presented. The abbreviations \textbf{SW}, \textbf{NX}, \textbf{RY}, and \textbf{GZ} in the environment names represent the size (\textbf{W}) of a room in the map, the number of obstacles (\textbf{X}), the number of rows (\textbf{Y}) and the number of goals (\textbf{Z}) during testing, respectively. The term \textit{``Close''} indicates that the agent requires door-opening actions.}}
    \label{tbl:performance_comparison}
    \vspace{-12pt}
\end{table}

\subsection{\textbf{Results}}
Table~\ref{tbl:performance_comparison} presents a summary of the simulation results. \model outperforms the baselines in \emph{trajectory planning}, achieving up to $25\%$ increase in success rates. It also has a larger margin over the baselines in challenging long-horizon tasks, demonstrating improved generalizability and adaptability. We also observe that the margin grows as the size of the environment increases. \model performs on par with LEAP on \emph{instruction completion} tasks while outperforming LEAP$\ominus$GC. It is also worth noting that LEAP outperforms other popular baselines, such as the Behavior Cloning algorithm (BC), and other model-free RL algorithms like Batch-Constrained deep Q-Learning~\cite{fujimoto2019off} and Implicit Q-Learning~\cite{kostrikov2021offline}, in single-goal tasks in BabyAI environments, as shown in Table 1 of LEAP~\cite{chen_planning_2023}. 

\noindent\textbf{Stochastic Environments for trajectory planning.}
Additionally, we evaluate trajectory planning in stochastic environments without any additional training. %
In this experiment, the agent has a 20\% chance of its chosen action \textit{left, right} being mapped to one of \textit{left, right, forward, pickup, drop, open}, with uniform probability. In a stochastic \textit{GotoLocal} environment, our model performs similar to how it did in the deterministic environment (table \ref{tbl:performance_comparison}). On the other hand, LEAP performs better in stochastic settings than it did in the deterministic environment. 
We observed that LEAP exhibits numerous stalling actions, such as \textit{in-place turns}. When these actions are mapped to other actions, there is a slight increase in success rates in smaller environments. However, a decline in performance is observed in larger mazes, as seen in the results for \textit{GotoObjMaze} in table \ref{tbl:performance_comparison}.

\noindent\textbf{Learning from random trajectories.}
\label{sec:random_result}
We assess the model's adaptive capabilities by training it on a single random-walk demonstration of a 100 time-steps.
Actions such as \textit{open}, \textit{pickup}, or \textit{drop} were absent from the demonstration. During the evaluation, the agent navigates a multi-room setting with doors as seen in \textit{MultiRoomN2S4}. \model outperforms LEAP 71\% to 0\% due to the stochastic nature of the policy and the diverse sub-goal distribution. On the other hand, LEAP gets stuck in a loop and never solves the task in any of the 150 runs.
\subsection{\textbf{Ablation Study}} 
\begin{table}[tb]
    \centering
    \small
    \setlength\tabcolsep{6pt} %
    \begin{tabular}{lcl}
        \toprule
        \textbf{Attribute} & \textbf{\model} &  \\
        \midrule
        \multicolumn{3}{c}{\textbf{Trajectory Planning - GoToObjMazeS4G2}} \\ 
        w/o GC & \textbf{32 $\pm$ 19\%} & \textcolor{red}{\textbf{$\downarrow$ 21 $\pm$ 3\%}} \\
        w/o action token + RTG & \textbf{25 $\pm$ 18\%} & \textcolor{red}{\textbf{$\downarrow$ 28 $\pm$ 2\%}} \\ 
        w/o entropy ($\tce$) & \textbf{33 $\pm$ 20\%} & \textcolor{red}{\textbf{$\downarrow$ 20 $\pm$ 4\%}} \\
        w/o discriminator ($\tdd, \tgen$) & \textbf{35 $\pm$ 24\%} & \textcolor{red}{\textbf{$\downarrow$ 18 $\pm$ 8\%}} \\
        \multicolumn{3}{c}{\textbf{Instruction Completion - GoToSeqS5R2}} \\ 
        w/o GC & \textbf{29 $\pm$ 2\%} & \textcolor{red}{\textbf{$\downarrow$ 16 $\pm$ 3\%}} \\
        w/o action token + RTG & \textbf{35 $\pm$ 3\%} & \textcolor{red}{\textbf{$\downarrow$ 10 $\pm$ 2\%}} \\ 
        w/o entropy ($\tce$) & \textbf{38 $\pm$ 2\%} & \textcolor{red}{\textbf{$\downarrow$ 7 $\pm$ 3\%}} \\
        w/o discriminator ($\tdd, \tgen$) & \textbf{38 $\pm$ 2\%} & \textcolor{red}{\textbf{$\downarrow$ 7 $\pm$ 3\%}} \\
        
        \bottomrule
    \end{tabular}
    \caption{{\small\textbf{Ablation.} Values correspond to the mean and variance of success rates as described in~\sect{sec:setup}. The numbers in red denote the decrease in success rates compared to those in table~\ref{tbl:performance_comparison}.}}
    \label{tbl:ablate}
    \vspace{-12pt}
\end{table}

\begin{figure}[tb]
    \centering
    \includegraphics[width=0.5\textwidth]{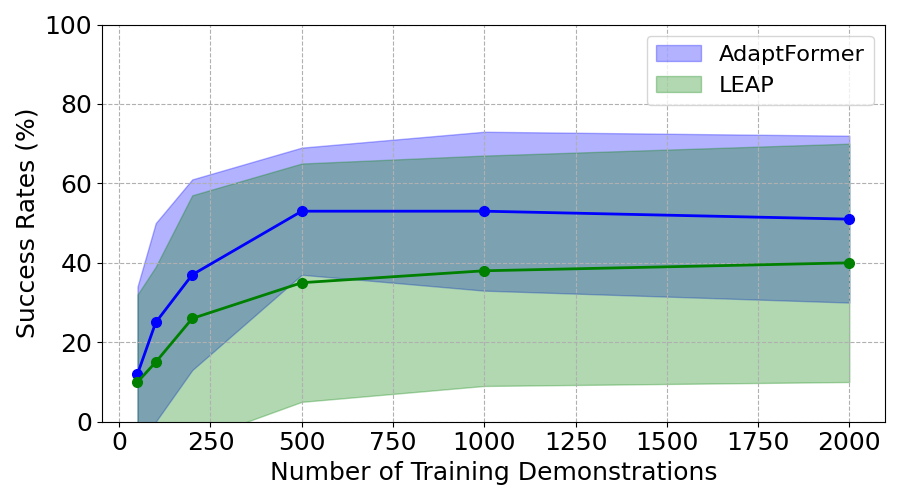}
    \caption{{\small\textbf{Number of Training Demonstrations vs. Success Rates.} We report the mean and standard deviation of success rates for the \textit{\textbf{GoToObjMazeS4G2}} task. Note that \model outperforms LEAP in mean success rates and shows lower variance.}}
    \label{fig:success}
    \vspace{-6pt}
\end{figure}

To assess the importance of the different components within \model, we perform a series of experiments by excluding each component in turn. Removing guidance from sub-goals (section~\ref{sec:ga}), the discriminator (equations~\ref{eq:disc_2} and~\ref{eq:gen}), or the entropy (term~\eqref{eq:ce}) significantly impacts the model's adaptability to test-time scenarios, leading to $\sim$10\% decrease in performance (see table \ref{tbl:ablate}). Qualitatively, excluding goal conditioning and the discriminator results in stalling actions (similar to LEAP). On the other hand, the absence of entropy regularization causes the agent to fail when goal-unblocking is required, as in Example~\ref{ex:adapt}. 
We also observed that by only learning state marginals (see line 2 in Table~\ref{tbl:ablate}), the model experiences a sharp drop in performance, as it fails to learn the transition dynamics.
\noindent\myparagraph{Effect of the Size of Training Data.} We evaluate success rates as a function of the number of training demonstrations, as shown in~\fig{fig:success}. There is a steep growth in success rates until 500 training demonstrations, after which learning saturates. %
It is important to note that our approach utilizes model-free \gls{rl} and does not explicitly learn dynamics. %
Moreover, due to the multi-goal setting, the model must generate trajectories that are approximately twice the duration of the training demonstrations. We observe that setting the planning horizon equal to the total sequence length improves scores for complex, long-horizon tasks (such as \textit{KeyCorridorS3R3}).
\vspace{-4mm}
\subsection{\textbf{Discussion of simulation results}}
\label{sec:discussion_results}
As seen in the presented results, \model outperforms the state-of-the-art across various tasks. In particular, it generalizes to unseen environments and tasks that were not seen during training. We briefly discuss some observations.

\noindent\textbf{Impact of goal conditioning.} Unlike Adaptformer, LEAP's goal conditioning considers only the final goal state as input, obtained through an oracle, which may not always be available. As seen via the ablation studies, goal conditioning, based on the intrinsic goal curriculum (see section \ref{sec:method}) %
is crucial for \model's adaptation to multiple goals. Without it, the agent tends to get stuck in a local region with no incentive to explore. %
In \model, we also iteratively re-initialize the sub-goals, preventing the agent from stalling in the same region, which helps it outperform the baselines on long-horizon tasks. %
Additionally, we observe that LEAP never takes goal-unblocking actions (such as moving an object in front of a door, see Example \ref{ex:adapt}), as these were not observed during training. Moreover, LEAP$\ominus$GC can only hallucinate trajectories from training and does not generalize. In contrast, our model demonstrates goal-unblocking capabilities and adapts to environments with closed doors, even without explicit demonstrations of door-opening actions.

\noindent\textbf{Correlation between energy and task.} 
\begin{figure}[tb]
    \centering
    \includegraphics[width = 0.5\textwidth]{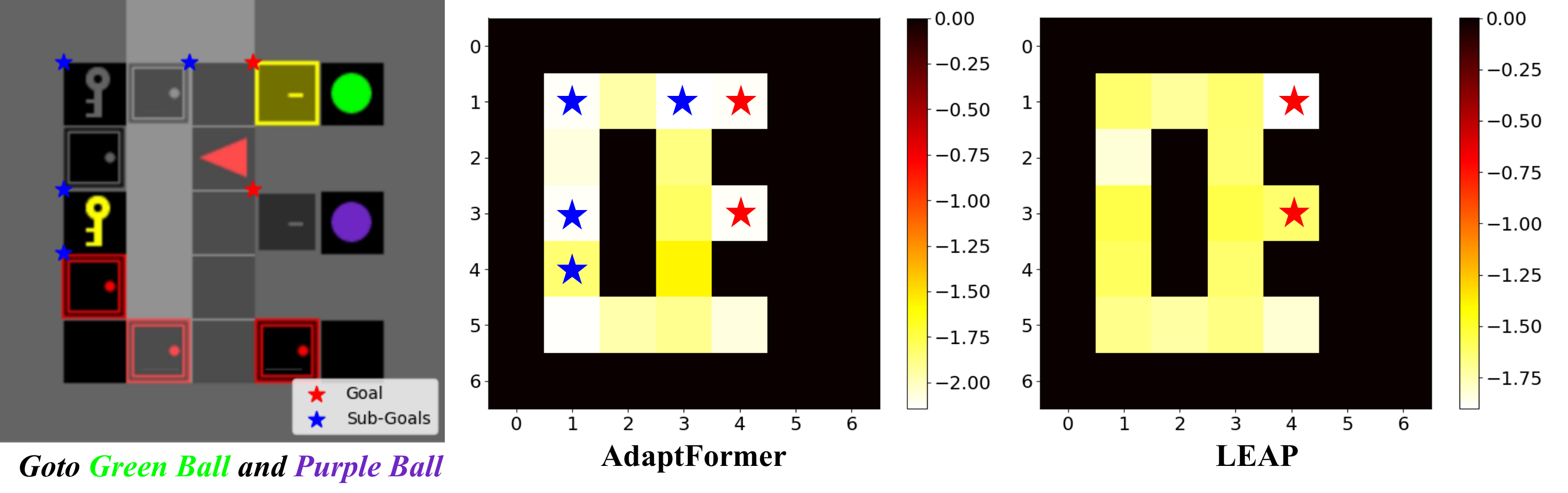}
    \caption{{\small\textbf{Energy Landscape.} AdaptFormer when conditioned with sub-goals, learns to implicitly assign minimum energy values to sub-goals (\textit{pick-up key, open doors}) required for task completion. States closer to the white region (low-energy) are more likely to be transitioned, indicating a higher probability of moving toward these preferred states. Conversely, LEAP does not pick up the sub-tasks associated with the task.}}
    \label{fig:enter-label}
    \vspace{-12pt}
\end{figure}

We explore the energy landscape and observe that states with low energy values are more likely to be visited. As illustrated in~\fig{fig:enter-label}, \model implicitly captures the sub-goals associated with the task, unlike LEAP, which conditions solely on the final goal state. %
The sub-goals effectively identify important state transitions, e.g., doors and keys, assigning them low energy. The iterative planner then estimates energy distributions following Algorithm~\ref{algo:iterative_plan} to generate a trajectory. %

\section{Conclusion}
\noindent\textbf{Summary.} We developed \model, a generative behavioral planner that can generalize to previously unseen tasks and environments. Extensive simulations demonstrate its capabilities, even in the presence of auxiliary distractors, and an improvement over the state-of-the-art. 

\noindent\textbf{Limitations.}
(1) We assume access to near-optimal demonstration trajectories for training; however, this might be unrealistic in some settings. Initial results show that \model performs well in simple tasks even when trained on random demonstrations. However, further studies are required to see how robust it is to sub-optimality in demonstrations.
(2) \model assumes complete information about a given environment. Initial experiments show that it can still complete simple tasks with only local information (such as objects in the agent's field of view); the method needs further development to work successfully in partially known environments. 
(3) While the distribution of $\Tau$ is stationary in the offline domain, the data distribution is non-stationary during the online adaptation. 

\noindent\textbf{Future work.} 
To overcome some of the limitations above, we will extend \model for online fine-tuning via hindsight experience replay \cite{zheng2022online, furuta2021generalized}. We will also further develop \model method to perform online information gathering in settings where the agent has a limited field of view.

\bibliographystyle{IEEEtran}  
\bibliography{references}  

\end{document}